\documentclass[10pt,twocolumn,letterpaper]{article}

\usepackage[pagenumbers]{cvpr} 

\usepackage[table]{xcolor}
\usepackage{cuted}
\usepackage{multirow}
\usepackage{longtable}
\usepackage{supertabular}

\newcommand{\ours}[0]{G-HOP}
\newcommand{\numcls}[0]{155}
\newcommand{\obman}[0]{3DW}

\definecolor{cvprblue}{rgb}{0.21,0.49,0.74}
\usepackage[pagebackref,breaklinks,colorlinks,citecolor=cvprblue]{hyperref}

\usepackage{bm}
\def\paperID{8666} 
\def\confName{CVPR}
\def\confYear{2024}

\title{\ours: Generative Hand-Object  Prior \\for  Interaction Reconstruction and Grasp Synthesis}

\author{
Yufei Ye\textsuperscript{1} \qquad Abhinav Gupta\textsuperscript{1} \qquad 
 Kris Kitani\textsuperscript{1,2}\qquad Shubham Tulsiani\textsuperscript{1}   \\
\textsuperscript{1}Carnegie Mellon University  \qquad \textsuperscript{2}MetaAI \\
\href{https://judyye.github.io/ghop-www/}{https://judyye.github.io/ghop-www}
}

\begin{document}
\maketitle

\begin{strip}
\vspace{-5em}
\centering
\includegraphics[width=\linewidth]{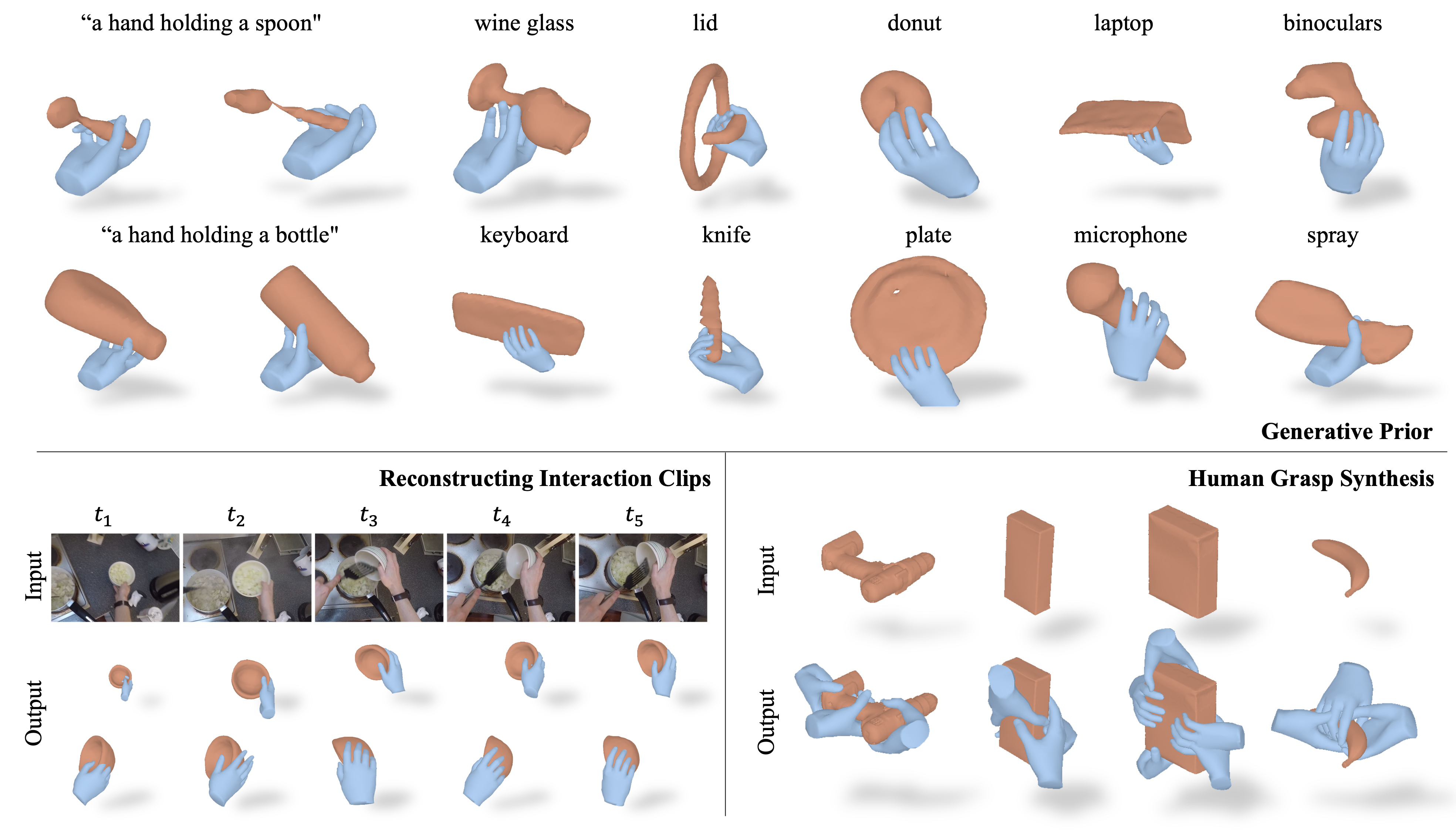}
\captionof{figure}{
\ours~can generate plausible hand-object interactions across a wide variety of objects (top). The learned generative prior can also guide inference for tasks such as reconstructing everyday interaction clips and synthesizing human grasps given object meshes. 
}
\label{fig:teaser}
\end{strip}


\begin{abstract}
\vspace{-1em}
We propose \ours, a denoising diffusion based generative prior for hand-object interactions that allows modeling both the 3D object and a human hand, conditioned on the object category. To learn a 3D spatial diffusion model that can capture this joint distribution, we represent the human hand via a skeletal distance field to obtain a representation aligned with the (latent) signed distance field for the object. We show that this hand-object prior can then serve as generic guidance to facilitate other tasks like reconstruction from interaction clip and human grasp synthesis. We believe that our model, trained by aggregating seven diverse real-world interaction datasets spanning across \numcls~categories, represents a first approach that allows jointly generating both hand and object. Our empirical evaluations demonstrate the benefit of this joint prior in video-based reconstruction and human grasp synthesis, outperforming current task-specific baselines. 

\end{abstract}    
\section{Introduction}
\label{sec:intro}

Imagine holding a bottle, or a knife, or a pair of scissors. Not only can you picture the differing shapes of these objects \eg a cylindrical bottle or a flat knife, but you can also easily envision the varying configurations your hand would adopt when interacting with each of them. Even though the form of these hand-object interactions  may vary widely depending on factors such as geometry  (\eg we will hold a pen and a pan rather differently), or intent (\eg passing a knife vs. using it to cut), we humans can effortlessly picture such interactions with everyday objects in our daily lives. In this work, our goal is to build a computational system that can similarly generate plausible hand-object configurations.

Specifically, we learn a denoising diffusion-based generative model that captures the joint distribution of both hand and object during interaction in 3D. Given  a category-conditioned description \eg `a hand holding a plate', our generative model can synthesize both, plausible object shape as well as the relative configuration and articulation of the human hand (see Fig. \ref{fig:teaser} top). A key question we address  is that what are good HOI \emph{representations} for the model. While objects shapes are typically described via spatial (signed) distance fields, human hands are commonly modeled via a parametric mesh controlled by an articulation variable. Instead of modeling these disparate representations in our generative model, we propose a homogeneous HOI representation and show that this allows learning a 3D diffusion model that jointly generates the hand and object.

In addition to enabling synthesis of diverse plausible hand and object shapes, our diffusion model can also serve as a generic prior to aid  inference across tasks where such a representation is a desired output.
For example, the ability to reconstruct or predict interactions is of central importance for robots aiming to learn from humans, or virtual assistant trying to aid them. 
We consider two well-studied tasks along these lines: i) reconstructing 3D hand-object shapes from everyday interaction clips, and ii) synthesizing plausible human grasps given an arbitrary object mesh.  To leverage the learned generative model as a prior for inference, we note that our diffusion model allows computing the (approximate) log-likelihood gradient given any hand-object configuration. We  incorporate this in an optimization framework that combines the prior likelihood-based guidance with task-specific objectives (\eg video reprojection error for reconstruction) or constraints (\eg known object mesh for synthesis) for inference.

While understanding hand-object interactions is an increasingly popular research area, real-world datasets capturing such interactions in 3D are still sparse. We therefore aggregate 7 diverse real-world interaction datasets resulting in long-tailed collection of interactions across 157 object categories, and train a shared model across these. To the best of our knowledge, our work represents the first such generative model that can jointly generate both, the hand and object, and we show that it allows synthesizing diverse hand-object interactions across categories. Moreover, we also empirically evaluate the prior-guided inference for the tasks of video-based reconstruction and human grasp synthesis, and find that our learned prior can help accomplish both these tasks, and even improve over task-specific state-of-the-art methods. 

\section{Related Works}
\label{sec:related_work}

\paragraph{Reconstructing Hand-Object Interactions.}
Reconstructing HOI interactions from images or videos can be challenging due to heavy mutual occlusions, and several initial approaches~\cite{hampali2020honnotate,brahmbhatt2020contactpose,garcia2018first,tekin2019h+} simplified the task by requiring an instance-specific object template and reducing the task to 6D pose estimation.
Some recent video-based reconstruction methods~\cite{huang2022hhor, wen2023bundlesdf,Hampali2022InHand3O} show promising results without requiring templates, but they target in-hand scanning setups where abundant multi-view cues are available and cannot infer unobserved regions.
Another line of template-free methods ~\cite{ye2022hand,chen2023gsdf,hasson19_obman,chen2022alignsdf, karunratanakul2020grasping,prakash2023learning} uses data-driven prior for reconstructing general objects from single images, but these are not temporally-consistent given input videos.  
Most closely related to our work is DiffHOI~\cite{ye2023vhoi} which leverages both multi-view cues and data-driven priors via per-sequence optimization. We adopt this framework and show the our proposed generative 3D prior can yield better reconstruction, while also enabling inference across other tasks. 

\vspace{-1.5em}\paragraph{Grasp Synthesis.}  Grasp synthesis studies how to interact with an objects plausibly. 
A line of work pursues 2D representations of interactions, or visual affordance. Given a 2D image, they predict interactions in various forms like trajectory, heatmaps, keypoints, or synthesized images~\cite{nagarajan2019grounded,fang2018demo2vec,ye2023affordance,liu2022joint}. However, interaction represented in 2D can not be directly used to command a robot to grasp an object in 3D.
There are extensive works in robotics that predict 3D robot grasp~\cite{li2007data,mahler2019learning,antotsiou2018task,brahmbhatt2019contactgrasp} for different end-effectors. Meanwhile, human grasp as a special end-effector receives great attention~\cite{karunratanakul2020grasping,jiang2021hand,elkoura2003handrix,kim2015physics,Grady2021ContactOptOC,brahmbhatt2020contactpose,liu2023contactgen}. Most relevant work including GF~\cite{karunratanakul2020grasping} and GraspTTA~\cite{jiang2021hand} model a conditional probability of human hand given an object mesh.  
In contrast to the task-specific methods, we directly leverage the generic joint hand-object generative prior and show that this leads to more natural human grasps.

\vspace{-1.5em}\paragraph{Diffusion Models as Generative Prior. } Diffusion models~\cite{ho2020denoising} are a family of generative models and have driven great progress in multiple domains like image generation~\cite{rombach2022high,ramesh2022hierarchical}, 3D object generation~\cite{jun2023shap,poole2022dreamfusion,lin2022magic3d}, novel-view synthesis~\cite{liu2023zero,melaskyriazi2023realfusion}, human motion~\cite{tevet2023human,karunratanakul2023guided}, video generation~\cite{singer2022make}, \etc.  An advantage of diffusion models is that they allow computing log-likelihood gradients via score distillation ~\cite{poole2022dreamfusion,wang2023score} and thus can be used as foundation generative priors for other tasks~\cite{zhou2023sparsefusion,melaskyriazi2023realfusion,Deng2022NeRDiSN,ruiz2023dreambooth}.
In our work, we use diffusion model to learn a generative prior for 3D hand-object interactions and apply it to the tasks of HOI reconstruction and grasp synthesis. 

\begin{figure*}
    \centering
    \includegraphics[width=\linewidth]
    {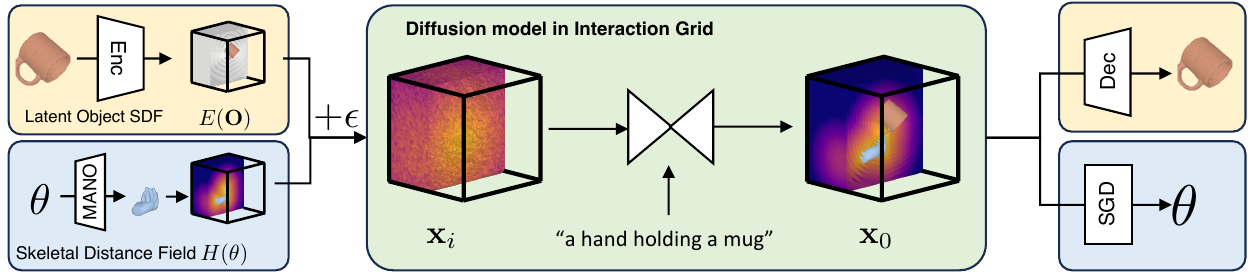}
    \vspace{-1.5em}
    \caption{\textbf{Method Overview of Generative Hand-Object Prior: } Hand-object interactions are represented as interaction grids within the diffusion model. This interaction grid concatenates the (latent) signed distance field for object  and skeletal distance field for the hand. Given a noisy interaction grid and a text prompt, our diffusion model predicts a denoised grid. To extract 3D shape of HOI from the interaction grid, we use decoder to decode object latent code and run gradient descent on hand field to extract hand pose parameters.}
    \vspace{-1em}
    \label{fig:method_prior}
\end{figure*}

\section{Method}
We first seek to model the joint distribution of the geometry of hand-object interactions $p(\mathbf O, \mathbf H | \mathbf C)$ where $\mathbf C$ is the text  of an object category.  We use a diffusion model $\Psi$ to learn this generative prior, and propose a spatial interaction grid representation for learning (Sec.~\ref{sec:prior}). 
We then apply this learned prior to guide reconstruction from monocular video clips and human grasp synthesis (Sec.~\ref{sec:app}). For both tasks, we frame inference as test-time optimization that combines task-specific constraints/objectives with score ``distillation" from the pre-trained diffusion model.

\subsection{Generative Hand-Object Prior}
\label{sec:prior}
In this work, we propose `interaction grids' as a homogeneous HOI representation that allows the diffusion models to effectively reason about the 3D hand-object interactions. 
Specifically, an interaction grid (Fig.~\ref{fig:method_prior}) is a concatenation of a latent signed distance value grid representing the object   $E(\mathbf{O})$ and a `skeletal distance' field based grid parameterized by 3D hand pose $H(\bm{\theta})$, \ie $\mathbf{x}\equiv (E(\mathbf{O}), H(\bm{\theta}))$.  
We model the interaction grid in a normalized hand-centric frame, where the hand palm always faces upwards. The hand-centric frame more effectively captures the inherent structures of interaction common to various objects, such as grasping handles, regardless of whether the object is a kettle or a power drill~\cite{ye2022hand}. 
\vspace{-1em}\paragraph{Latent Object Signed Distance Field. } We use a signed distance field (SDF) grid to capture object details.  As the memory grows cubically with grid resolution,  we follow prior works to use a VQ-VAE~\cite{van2017neural} to compress high-resolution SDF grids into lower-dimension object latent. $\mathbf z = E(\mathbf O), \mathbf O = D(\mathbf z)$. Note that when training the autoencoder, the object SDF grids are also transformed into hand-centric frame. 

\vspace{-1em}\paragraph{Skeletal distance field for Parametric Hand.}  While there is consensus on how to represent objects, it is unclear what is a good representation of hand during interaction. 
Many prior works generate hand/human shape by diffusing in the compact pose parameter space~\cite{tevet2023human,karunratanakul2023guided}  but we find this space not ideal when we diffuse it jointly with objects latent grids (see supplementary) probably because the diffusion model cannot easily to reason about spatial interactions using this heterogeneous representation (1D articulation vector and 3D SDF grid).  Instead, we propose to represent hand in a pose-parameterized distance field $H(\bm \theta)$. It is a 15-channel 3D grid that encodes the distance to each joint. $H(\bm \theta)[u, v, w]_{i=1:15} \equiv \|\mathbf X_{[u,v,w]} - J_i\|_2^2$. 
This skeletal distance field can be converted from pose parameter space and vice versa by leveraging differentiable parametric mesh model MANO~\cite{mano}.  
Specifically, MANO takes in the pose parameter and outputs joint position $J_i(\bm \theta)$ to compute the skeletal field. 
To recover pose parameter $\bm \theta$ from a skeletal distance field, we run gradient decent on pose parameter to minimize the distance between the  induced field and the given field, 
$\bm \theta^* = \arg \min_{\bm \theta} (H(\bm \theta) - \hat H) + w \|\bm \theta \|_2^2$.

\vspace{-1em}\paragraph{Denoising Diffusion Model.}  
In training, the diffusion model takes in a text embedding and a noisy 3D interaction grid $\mathbf{x}_i$ and is supervised to restore the clean grid $\hat{\mathbf{x}}_0$.
\begin{align}
    \mathcal{L}_{\text{DDPM}}[\mathbf{x}; \mathbf{C}] = \mathbb{E}_{i, \epsilon \sim \mathcal{N}(\mathbf{0}, \mathbf{I})} w_i \| \hat{\mathbf{x}}_0- \Psi(\mathbf{x}_i, i, \mathbf{C}) \|^2_2
\end{align}
The object distance field is in resolution $64^3$  and the VQ-VAE downsamples the resolution to $16^3$ which is then concatenated with the hand skeletal field. We implement the diffusion model as 3D-UNet with three 3D convolution blocks. The text prompt is encoded by CLIP~\cite{CLIP} text encoder and is passed to the 3D-UNet by cross-attention at each block. 

\subsection{Prior-guided Reconstruction and Generation}
\label{sec:app}
Given the learned generative prior,  we leverage it for both HOI reconstruction and human grasp synthesis.  
The inference in both tasks is performed via test-time optimization which is guided by distilling the learned prior.  
We use score distillation sampling (SDS~\cite{poole2022dreamfusion,wang2023score}) to approximate log-probability gradients of interaction grids $\mathbf x$ from the diffusion model. Specifically, to guide the grid to be more plausible at every optimization step, we corrupt the current interaction grid  $\mathbf x$ by a certain amount of noise and let diffusion model denoise it. The discrepancy between this denoised prediction and the current estimate can be be used an objective to obtain log-likelihood gradients: 
\begin{align}
    \nabla_{\mathbf{x}} \log p(\mathbf x) \approx \mathcal \nabla_{\mathbf{x}} L_{SDS} [\mathbf{x}] = \mathbb E_{\epsilon, i} 
    [w_i (\mathbf x - \hat{\mathbf{x}}_i)]
\label{eq:sds}
\end{align}

In the following section, we will show that both reconstruction and grasp synthesis can leverage the common optimization frameworks by instantiating task-specific parameters and constraints.

\begin{figure}
    \centering
\includegraphics[width=\linewidth]{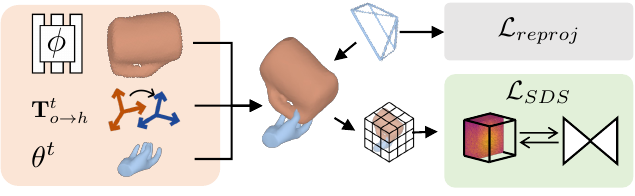}
\caption{\textbf{Reconstructing  Interaction Clips:} We parameterize HOI scene as object implicit field, hand pose, and their relative transformation (left).  The scene parameters are optimized with respect to the SDS loss on extracted interaction grid and reprojection loss (right). }    \label{fig:method_recon}
\vspace{-1em}
\end{figure}

\subsubsection{Reconstructing Interaction Clips}
\label{sec:recon_video}
Given a video clip depicting a hand interacting with a rigid object, we aim to reconstruct the underlying 3D shape of the hand and the object. We follow DiffHOI~\cite{ye2023vhoi}  which performs inference via a optimizing 3D scene representation with respect to a reprojection term and a data-driven prior term. Instead of their 2D diffusion prior which can only guide object shape inference, we substitute our learned joint 3D generative prior and show that it leads to improved performance for video-based reconstruction.

\vspace{-1em}\paragraph{Scene Parameters and Rendering. } We adopt a similar representation as DiffHOI~\cite{ye2023vhoi}, which decomposes the HOI scene into three parts: i) a time-persistent object signed distance field represented by an implicit neural network $\phi(\cdot)$; ii) time-varying hand pose parameters $\bm \theta^t$, and  iii) the relative poses $\mathbf T_{o\to h}^t$ between them.  
This scene representation can be rendered into 2D masks $\mathbf I^t$ by differentiably compositing renderings of the volumetric object and hand mesh.

\vspace{-1em}\paragraph{Prior-Guided Reconstruction.}  
Different from DiffHOI, our data-driven prior is in 3D space instead of 2D. Furthermore, our prior also models the hand pose rather than use it as a condition, and can thus also provide gradients to guide hand pose optimization. Specifically, to regularize the 3D representation, we query the 3D volume in the hand-centric frame to get interaction grid for each frame and pass the grid to the pre-trained diffusion model, \ie
$\mathbf{x}^t = (E(\phi(\mathbf T^{t^{-1}}_{o\to h} X_{grid})), H(\bm \theta^t) )$,
where $X_{grid}$ is the coordinate of the queried volume. 
Other losses are similar to~\cite{ye2023vhoi}: 
the reprojection term is computed in the mask space $\mathcal{L}_{reproj} = \|\mathbf I^t - \hat{\mathbf{I}^t}\|$; other regularization include Eikonal loss and temporal smoothness.

The optimization converges faster than previous work, perhaps because the prior in 3D provides stronger supervision. Specifically, we optimize 15000 iterations for each video clips which takes about an hour (which is ~85\% faster than DiffHOI~\cite{ye2023vhoi}).

\begin{figure}
    \centering
    \includegraphics[width=\linewidth]{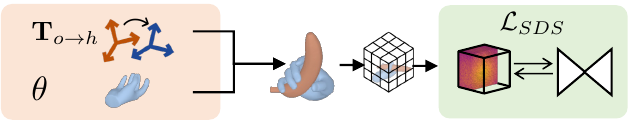}
    \vspace{-2em}
    \caption{\textbf{Grasp Synthesis:} We parameterize human grasps via hand articulation parameters and the relative hand-object transformation (left).  These are optimized with respect to SDS loss by converting grasp (and known shape) to interaction grid (right). }
    \label{fig:method_grasp}
\vspace{-1em}    
\end{figure}

\begin{figure*}
    \centering
    \includegraphics[width=\linewidth]{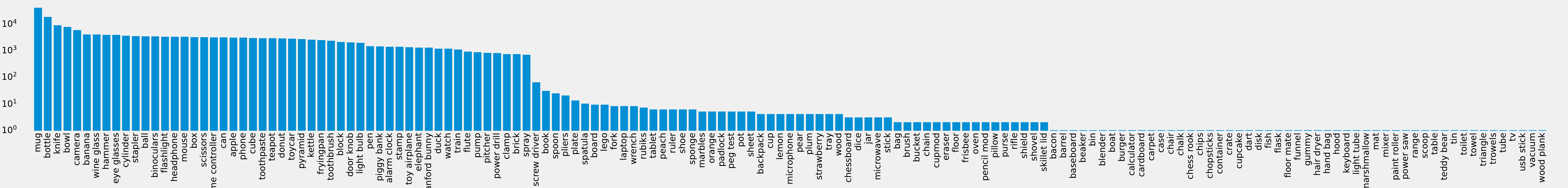}
    \caption{\textbf{Dataset Statistics: } number of training samples for each category when training our generative prior. Zoom in for better view. }
    \label{fig:data}
    \vspace{-1em}
\end{figure*}
\begin{figure*}
    \centering
    \includegraphics[width=\linewidth]{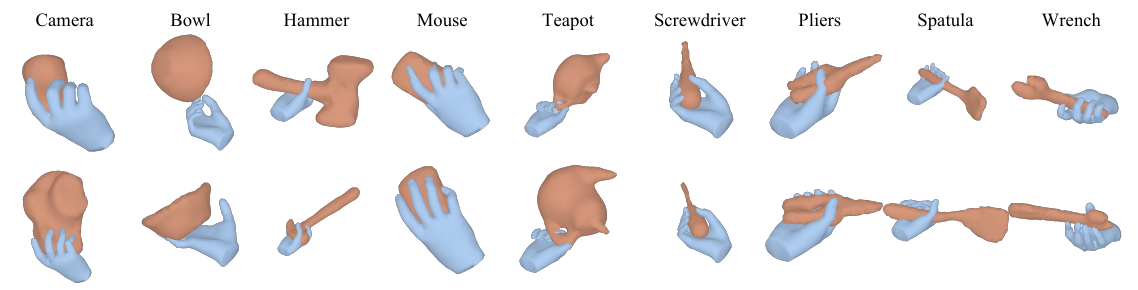}
    \vspace{-1.5em}
    \caption{\textbf{Generations from Generative Hand-Object Prior: } Given a text prompt (only showing class label), we visualize two generated interactions from \ours~.  Categories are sorted from most common to least common in training (left to right). Generations are diverse in terms of object shape such as teapots, hand articulation such as mouse, and use intent like hammer. }
    \label{fig:prior}
    \vspace{-1em}
\end{figure*}

\subsubsection{Synthesizing Plausible Human Grasps}
\label{sec:grasp}
Given an object mesh $M_o$, we aim to synthesize human grasps for the  object. Formally, this corresponds to sampling from the conditional distribution $p(\mathbf H|\mathbf O, \mathbf C)$.  While our diffusion model captures the joint distribution of hand and object, it does not allow sampling human grasp directly given an object. Instead, we obtain plausible grasps via a test-time optimization approach to seek grasping modes while constraining the object to match the input. We also provide a mechanism to rank the generation by measuring consensus between diffusion model and the grasp synthesis.

\vspace{-1em}\paragraph{Grasp Parameters. } We parameterize a human grasp  by the relative pose of the hand with respect to the object $\mathbf T_{o\to h}$, along with its articulation $\bm \theta$. 
We initialize hand articulation to a mean configuration while initializing relative pose with a random orientation and translation.

\vspace{-1em}\paragraph{Optimization. } In order to use diffusion model to guide grasp synthesis, we first convert the object mesh into SDF grid $G_o$, which is then transformed from the object-centric to the diffusion model coordinate (hand-centric) by the relative pose $\mathbf T_{o\to h}$, \ie
$\mathbf x = (E(\mathbf T_{o\to h} G_o), H(\bm \theta))$.    
We optimize the relative pose along with hand articulation for 500 iterations by maximizing the interaction likelihood from Eq.~\ref{eq:sds}, \ie $\log p(\bm x (\bm T_{o\to h}, \bm \theta))$. 
To account for accuracy loss when converted to low-resolution grids, we refine the predicted hand with the original mesh to encourage surface contact and penalize mesh collision. 
We show in supplementary that the distillation provides a good initialization for the mesh refinement while surface refinement further improves contact and grasp stability.

\vspace{-1em}\paragraph{Ranking Grasps. }
The proposed approach to grasp synthesis is stochastic due to different initialization and the stochastic distillation process. Thus diverse grasps can be sampled. Furthermore, many applications like robotic manipulation would also want to know how plausible each grasp is.  We also propose a mechanism to evaluate the sampled grasp. We approximate the likelihood upper bound~\cite{Ho2020DenoisingDP} by averaging SDS loss across different time steps $i$:
\begin{align}
    s(\mathbf \bm \theta, \mathbf{T}_{o\to h}) = -\sum_{i=1}^T w_i \|\mathbf{x}(\bm \theta, \mathbf{T}_{o\to h}) ) - \hat{\mathbf x}_i(\epsilon)\|_2^2
\end{align}
Intuitively, this measures the agreement between the prediction and the denoised output from the diffusion model, which indicates the distance of the current grasp to a plausible mode.  We observe that this score provides a consistent and meaningful ranking across different samples.

\section{Experiments}
\label{sec:exp}

We train the generative prior on a collection of HOI datasets. We first show data distribution on this dataset collection and then visualize samples from the learned generative prior (Sec~\ref{sec:exp_prior}). In Sec.~\ref{sec:exp_recon}, we show that the learned prior benefits the task of reconstructing interaction clips.  Our method outperforms other reconstruction baselines on HOI4D and we also show reconstruction of in-the-wild videos. In Sec.~\ref{sec:exp_grasp}, we evaluate human grasps that are synthesized by directly applying  our learned prior. We compare \ours~ to other baselines on two datasets and conduct user study to show that human grasp synthesized by ours is the most preferred one.  

\begin{figure*}
    \centering
    \includegraphics[width=\linewidth]{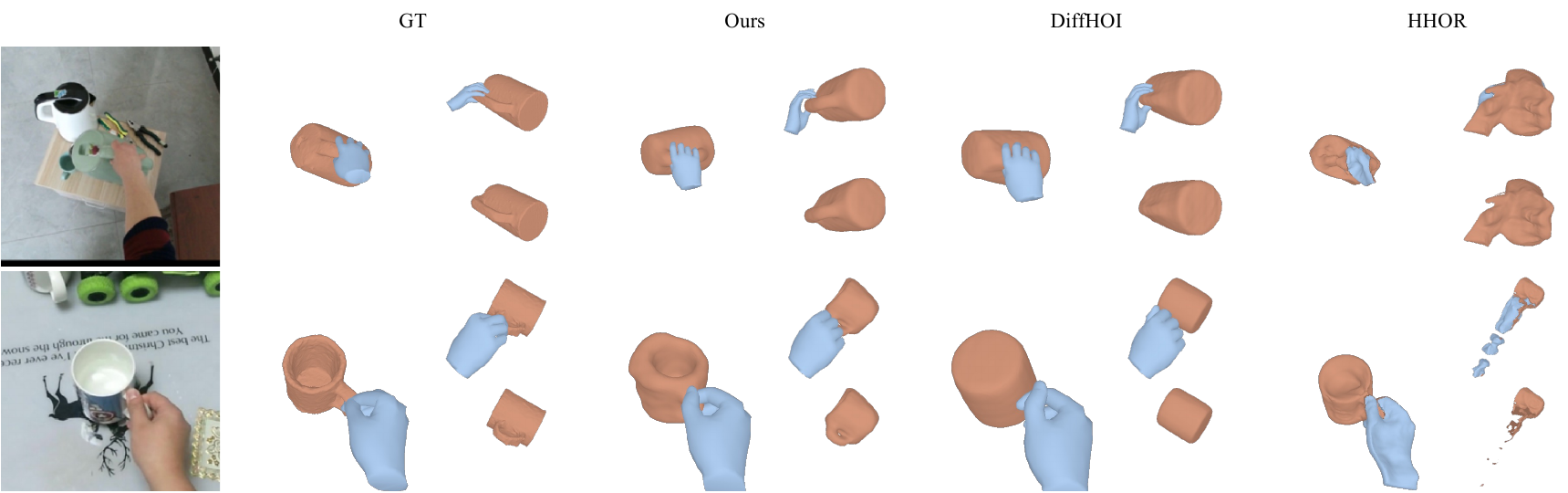}
    \caption{\textbf{Qualitative Evaluation on HOI4D:} We show reconstruction by \ours~and two other video reconstruction baselines~\cite{ye2023vhoi,huang2022hhor} in the image frame (left) and from another view with (top right) or without (bottom right) reconstructed hand. Please see our project page for reconstruction videos from all methods. }
    \label{fig:recon}
    \vspace{-1em}
\end{figure*}
\vspace{-1.2em}\paragraph{Training Data. }    We train our diffusion model on a combination of several world datasets including~\cite{hoi4d,YangCVPR2022OakInk,taheri2020grab,brahmbhatt2020contactpose,corona2020ganhand,dexycb}, using their annotated 3D meshes of hand and objects. 
The name of categories across datasets are not standardized so we manually map synonyms or different formats to the same word (\eg cellphone, iphone $\to$ phone, doorknob, door\_knob $\to$ door knob). In total, we reduce 362 different words to \numcls~classes. 
All training data were converted into SDF grids, in hand-centric frame, with a resolution of $64^3$ and spanning 30cm in all directions.

\subsection{Visualizing Data-Driven Prior}
\label{sec:exp_prior}
We visualize the number of training samples per class in Fig.~\ref{fig:data}. The data is extremely unbalanced and follows a long-tail distribution. Classes with most training samples like mug consist of more than 10k grasps while few-shot classes such as skillet lid consist of fewer than 100 grasps. 

In Fig.~\ref{fig:prior}, we visualize hand-object interactions generated from the learned generative prior. We show 3 samples in different rows for each class.  The classes from left to right are sorted by the training size from more to less. 
We see that the generated objects vary in shape. For example, different cameras display various lengths of lens. The generated samples are also diverse in terms of ways to hold them. Some hammers are held by handles and some are held by heads (for hand-over). We also find that the model can generate diverse and plausible samples on few-shot classes (shown on the right side).

\subsection{Reconstructing Interaction clips}
\label{sec:exp_recon}
\paragraph{Setup and Evaluation Metrics. } We evaluate interaction reconstruction on the HOI4D dataset. HOI4D is an egocentric dataset recording people interacting with different objects. We use the same split as DiffHOI~\cite{ye2023vhoi} that consists of 2 video clips for all portable rigid object categories. The objects in the test set are held out from the train set. 
We evaluate three aspects of the output: object reconstruction error, hand reconstruction error (MPJPE, AUC), and hand-object alignment (CD$_h$). Following prior works~\cite{ye2023vhoi,huang2022hhor}, we align the object reconstruction with the ground truth by scaled Iterative Closest Points (ICP) and report F-score at 5mm, 10mm, and Chamfer distance in the aligned space. 
To evaluate the relation between hand and object, we report Chamfer distance of objects in hand-centric frame $CD_h \equiv CD(\mathbf T_{o\to h}^t O, \hat{\mathbf{T}}^t_{o\to h}\hat O)$.

\begin{figure}
    \centering
    \includegraphics[width=\linewidth]{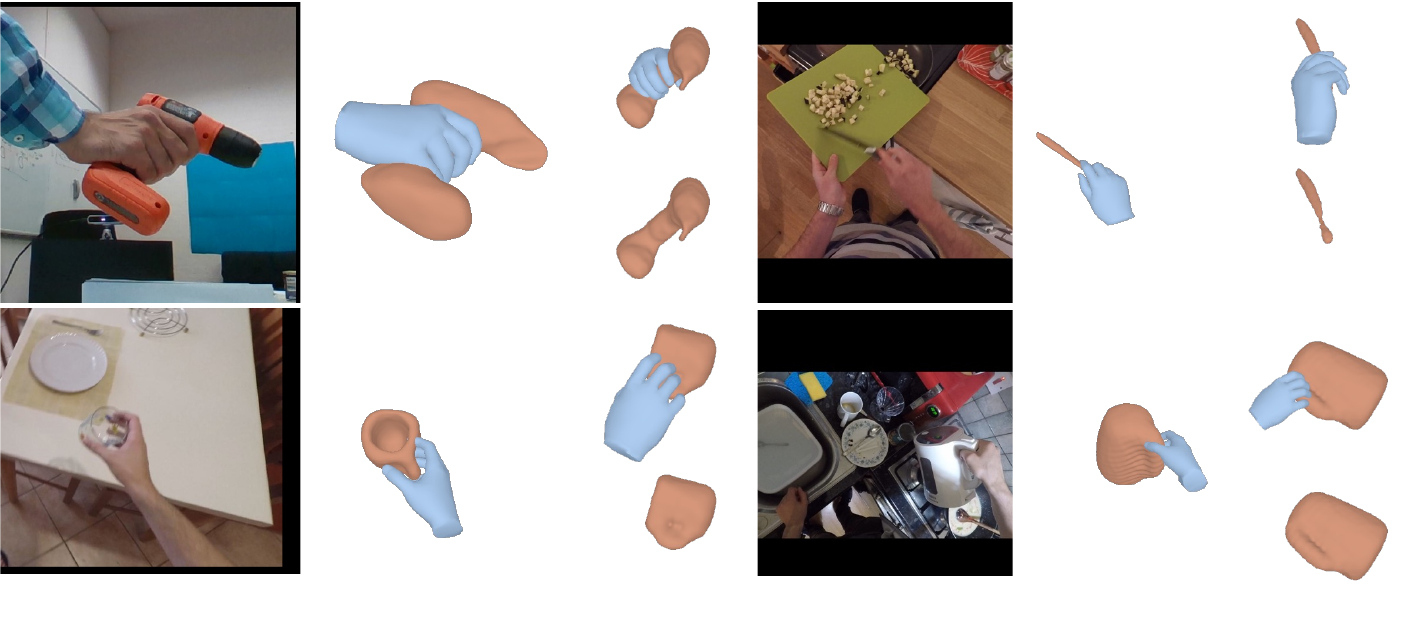}
    \vspace{-1.5em}
    \caption{\textbf{In-the-Wild Reconstruction: } reconstruction on interaction clips  from novel datasets  \cite{Damen2018EPICKITCHENS,hampali2020honnotate}.} 
    \label{fig:ood_recon}
\vspace{-1em}    
\end{figure}
\begin{figure*}
    \centering
    \includegraphics[width=\linewidth]{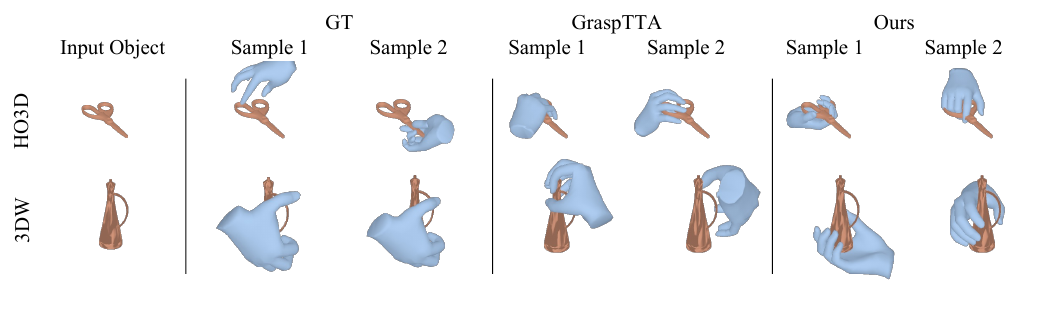}
    \vspace{-3.5em}
    \caption{\textbf{Visualizing Grasp Generations:} Given an object mesh (left) from HO3D or \obman, we sample two grasps from each method. 
    }
    \label{fig:grasp}
    \vspace{-1em}
\end{figure*}

\vspace{-1.2em}\paragraph{Baselines. }
We compare with three other template-free baselines that tackle reconstruction from casual monocular interaction clips.
\\i) \textit{iHOI}~\cite{ye2022hand} is a single-view 3D reconstruction method that learns to map from image feature and hand articulation to in-hand object shape. The model is finetuned on HOI4D and reconstruction is evaluated per-video frame. 
\\ii) \textit{HHOR}~\cite{huang2022hhor} optimizes a hand-object field with respect to the input video without any data-driven prior. 
\\iii) \textit{DiffHOI}~\cite{ye2023vhoi} is  closest to our work. The main difference is that the prior in their work takes hand pose as input thus modeling the \textit{conditonal} probability $p(\pi(O) | \pi (H), C)$.  Additionally, their prior is an image-based diffusion model instead of a 3D diffusion model. 
\\iv) \textit{\ours~(Cond)} is our ablated models that is conditioned on hand pose and text prompt (same as DiffHOI but with 3D backbone). It aims to disentangle the effect of upgrading the prior from 2D to 3D from modeling joint instead of conditional probability. 
\\For fair comparison, our diffusion model for HOI4D evaluation only trains on HOI4D train split.  All other experiments use the model trained on all datasets.  


\definecolor{first}{rgb}{1.0, .83, 0.3}
\definecolor{second}{rgb}{1.0, 0.93, 0.7}
\def \first {\cellcolor{first}}
\def \second {\cellcolor{second}}
\def \third {}

\begin{table}[t]
\footnotesize
\begin{center}
\vspace{-1em}
\caption{\textbf{Comparing HOI reconstruction:} object error (F@5mm, F@10mm, CD), hand-object alignment CD$_h$, and hand error (MPJPE, AUC) on HOI4D. We compare \ours~ with baselines and also ablate if reconstruction benefits from priors in the 3D space or from joint modeling hand and object. }
\vspace{-1em}
\label{tab:recon}

\setlength{\tabcolsep}{2pt}
\resizebox{\linewidth}{!}{
\begin{tabular}{l c c c c c c c}
\toprule

& \multicolumn{3}{c}{Object Error} 
& \multicolumn{1}{c}{Align}
& \multicolumn{2}{c}{Hand Error} \\

\cmidrule(r){2-4} \cmidrule(r){5-5} \cmidrule(r){6-7}

& F5$\uparrow$  & F10$\uparrow$ & CD$\downarrow$  & CD$_h$ $\downarrow$ & MPJPE$\downarrow$ & AUC$\uparrow$ \\

\midrule
iHOI~\cite{ye2022hand} & 0.42 & 0.70 & 2.7 &  27.1 & {1.19} & 0.76 \\
HHOR~\cite{huang2022hhor} & 0.31 & 0.55 & 4.7 & 165.4 & - & - \\
DiffHOI~\cite{ye2023vhoi} &  0.62 &  0.91 &  0.8 & 48.7 &  1.12 &  0.78 \\
\ours &  \textbf{0.76} &  \textbf{0.97} &  \textbf{0.4} &  \textbf{18.4} &  \textbf{1.05} &  \textbf{0.79} \\
\midrule
\ours (Cond) & 0.66 & 0.92 & 0.7 &  19.3 & 1.14 & 0.77 \\
\bottomrule
\end{tabular}
}
\end{center}
\vspace{-2em}
\end{table}

\vspace{-1.2em}\paragraph{Results. } We visualize reconstructions from different methods in Fig.~\ref{fig:recon} in the image frame and from a novel viewpoint. HHOR, which does not leverage data-driven learning, struggles with unobserved regions and outputs degenerate solutions as shown from the novel view. While iHOI reconstructs better shapes for each frame, there are not temporally consistent (shown in supplementary video) and it cannot benefit from multi-view cues. In comparison, DiffHOI reconstructs temporally consistent and more realistic results, but the reconstructed shape is relatively coarse. For instance, the kettle handle is merely a bump on top of a cylinder and the reconstruction does not reflect the concavity of the mug. In contrast, the reconstruction from \ours~captures more details of object shape. In the bottom row, it even captures the space between the handle and the cup body. The visualization is consistent with the quantitative results in Tab.~\ref{tab:recon}. Furthermore, we also find that the hand pose reconstruction also improves since the prior in \ours~can also guide hand pose as well. 

\vspace{-1.2em}\paragraph{Ablations.} Comparing with the ablated 3D conditional model (Tab.~\ref{tab:recon}), we find that upgrading 2D prior to 3D improves object reconstruction significantly but does not improve hand reconstruction much. Joint modeling leads to better hand pose, which can in return improve object reconstruction further. 
Interestingly, we also find that the variant that \textit{jointly} models HOI in \textit{image} space performs even worse than DiffHOI. See appendix (2D joint prior) for further discussion.

\begin{figure}
    \centering
    \includegraphics[width=\linewidth]{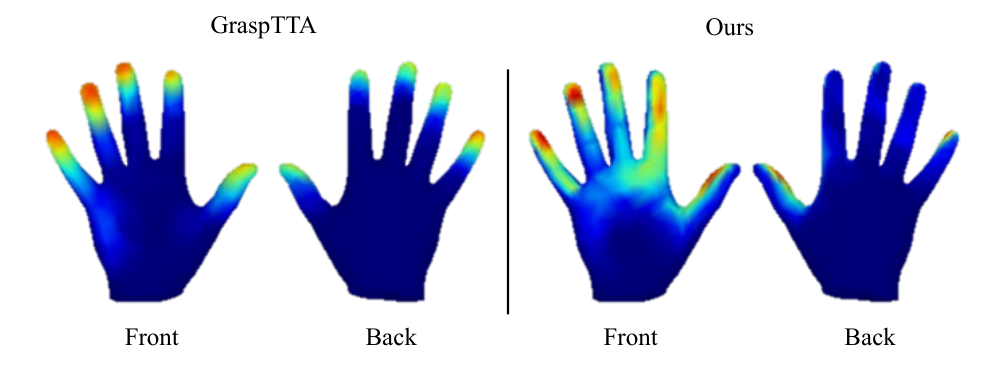}
    \vspace{-1.5em}
    \caption{\textbf{Contact Map on Hand:} We visualize contact probability on hand over all generated samples from \ours~ and GraspTTA~\cite{jiang2021hand} on the HO3D dataset. 
    }
    \label{fig:contactmap}
    \vspace{-1.5em}
\end{figure}

\begin{figure}[ht]
    \centering
    \begin{minipage}{\linewidth}
        \centering
        \begin{center}
\captionof{table}{\textbf{Comparison with Baselines:} We compare our synthesised human grasps against GraspTTA~\cite{jiang2021hand} and annotated grasps provided by datasets (GT) on HO3D and \obman. We report table the intersection between meshes, displacement distance in simulation, and hand contact ratio and area  (top).  We also report preference percentages from users for pairwise method comparison on HO3D and \obman~(bottom). 
}
\vspace{-1em}
\label{tab:grasp}

\setlength{\tabcolsep}{2.7pt}
\resizebox{1\linewidth}{!}{
\begin{tabular}{l l c c c c c c c c }
\toprule

 &
& \multicolumn{3}{c}{Intersection} 
& \multicolumn{1}{c}{Disp.}
& \multicolumn{2}{c}{Contact}
\\
\cmidrule(lr){3-5} \cmidrule(lr){6-6} \cmidrule(lr){7-8}

& & maxD$\downarrow$  & avgD$\downarrow$ & vol$\downarrow$ & avg $\downarrow$  & ratio$\uparrow$ & area$\uparrow$ \\
\midrule
\multirow{3}{*}{\rotatebox{90}{HO3D}}
& GT &  \textbf{1.32} &  0.37 &  6.16 &  2.32 & 0.95 & 0.15 \\
& GraspTTA & 2.44 & 0.61 &  \textbf{5.25} & 2.89 &  1.00 &  0.23 \\
& \ours &  1.84 &  \textbf{0.31} & 11.46 &  \textbf{0.95} &  \textbf{1.00} &  \textbf{0.23} \\

\midrule

\multirow{3}{*}{\rotatebox{90}{\obman}} 
& GT* & 0.98 & 0.74 &  \textbf{1.70} &  1.57 &  1.00 & 0.12 \\
& GraspTTA &  0.87 &  0.58 &  5.56 &  \textbf{1.54} &  \textbf{1.00} &  0.18 \\
& \ours &  \textbf{0.74} &  \textbf{0.51} & 17.40 & 1.85 & 0.93 &  \textbf{0.25} \\

\bottomrule
\end{tabular}
}

\end{center}
    \vspace{-0.5em}
    \end{minipage}\hfill

    \begin{minipage}{\linewidth}
        \centering
        \includegraphics[width=\linewidth]{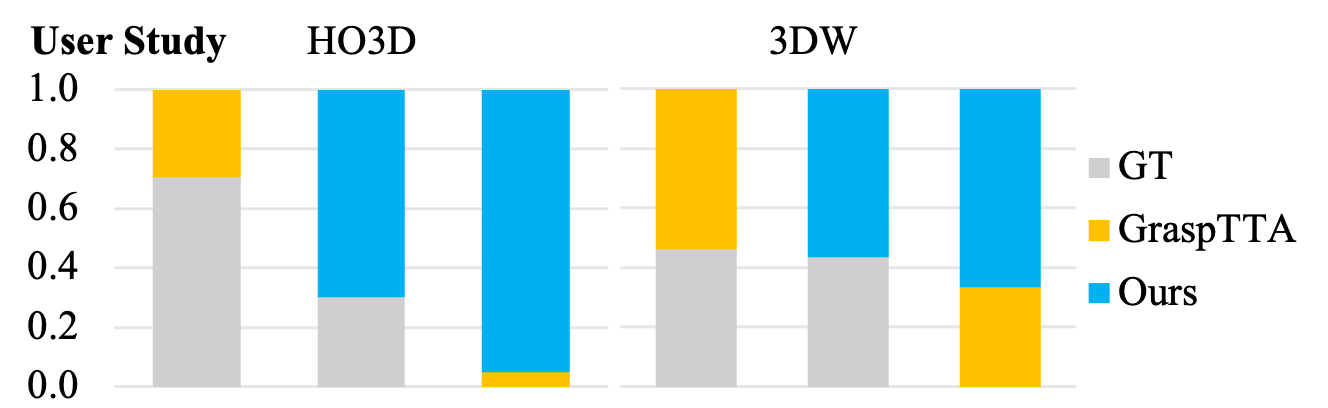}
    \end{minipage}\hfill
    
    \vspace{-1em}
\end{figure}

\subsection{Synthesizing Plausible Grasps}
\label{sec:exp_grasp}
\paragraph{Setup and Evaluation  Metrics. } We evaluate human grasp synthesis on two datasets~\cite{hampali2020honnotate,hasson19_obman}. HO3D is a real-world HOI dataset whose objects come from the YCB dataset~\cite{calli2015ycb}, which has appeared in our training data. To test the generalization ability to novel objects, we also evaluate on a subset of 3D Warehouse used in Hasson \etal~\cite{hasson19_obman} (\obman). It is a synthetic dataset that our prior has never seen in training. 
Following prior work~\cite{karunratanakul2020grasping, jiang2021hand}, we evaluate grasp quality by 1) the amount of intersection between hands and objects (mean volume, maximum and mean depth), 2) the displacement of objects when placed into simulation~\cite{coumans2021}, and 3) the contact hand region (ratio and area, where ratio is the percentage of grasps that have non-zero contact area). 
There is a trade-off between contact/simulation displacement and intersection. While the metrics characterize the grasp quality, no single metric alone is conclusive on grasp synthesis.   So we also conducted a user study. We show users two human grasps randomly chosen from two methods and ask them to select their preferred one. We collected 440 and 380 answers from 22/19 users on HO3D and \obman~  accordingly.

\vspace{-1.2em}\paragraph{Baselines. }
We compare with baseline GraspTTA~\cite{jiang2021hand} which is trained on in-domain data (\obman~with annotated grasps). It learns to generate contact maps on hand and object which are then optimized along with hand pose be self-consistent during test time. We also compare with ground truth annotation in both datasets. While Grasping Fields~\cite{karunratanakul2020grasping} is also a representative method for grasp generation, their evaluation setup assumes a known object pose relative to the hand unlike ours, and randomizing this relative pose significantly affects their performance. We detail this further and report our results under their evaluation setting in supplementary.

\vspace{-1.2em}\paragraph{Results. }
Fig.~\ref{fig:grasp} visualizes two human grasp synthesis from each method for a given object. Annotated grasps (GT) in two datasets display different grasping styles. 
Semi-automatically generated grasps~\cite{hasson19_obman} sometimes do not look natural and tend to ``over-grasp" as they are generated to maximize stability. 
GraspTTA is trained on the same dataset and shows similar over-grasp behavior while our grasps appear more natural. 
In contrast, \ours~ grasps objects from different directions while all of the synthesized hands make contact with the objects. 

\begin{figure}[t]
    \centering
    \includegraphics[width=0.95\linewidth]{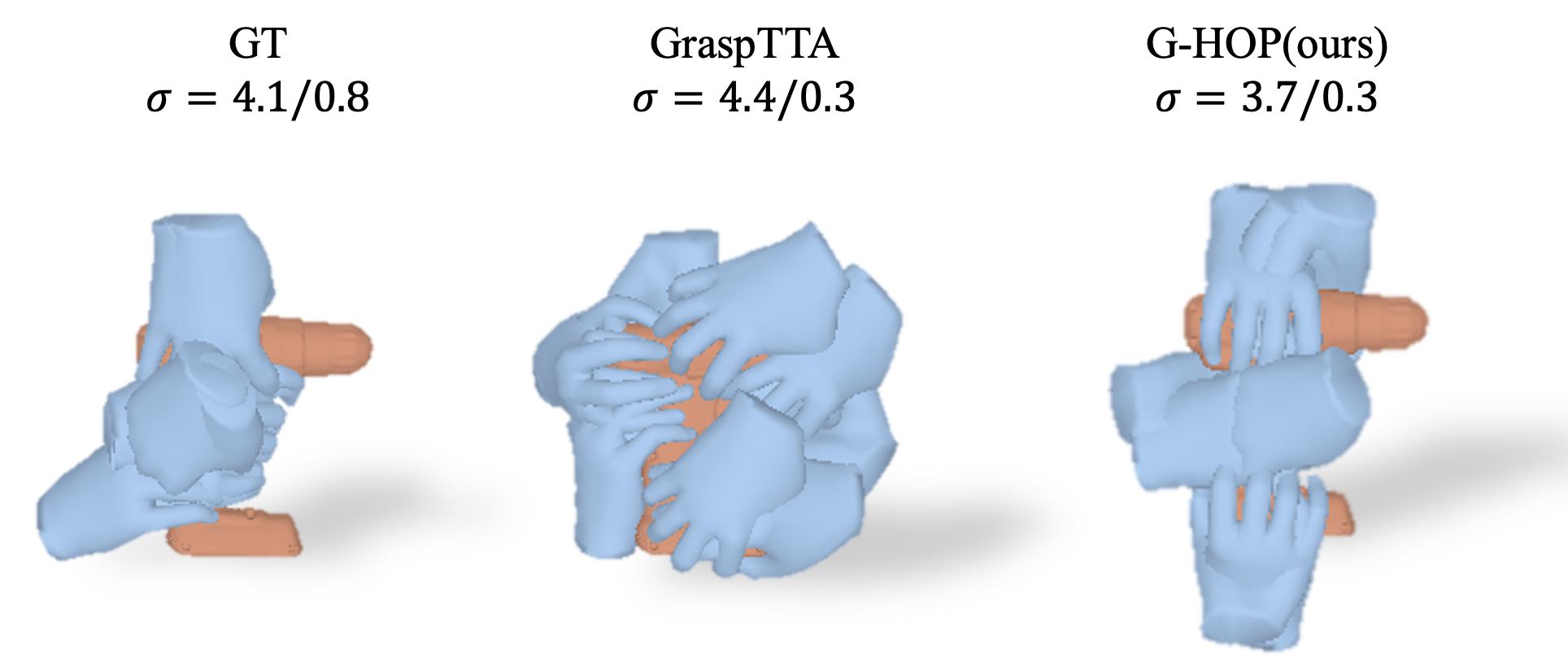}
    \vspace{-0.3cm}
    \caption{\textbf{Grasp Diversity:} 10 random grasps of a power drill.  
    Although GraspTTA generates more diverse grasps, some of them are not plausible as they disregard object functions.
    }
    \vspace{-1.5em}    
    \label{fig:diversity}
\end{figure}

\vspace{-1.2em}\paragraph{Grasp Diversity. }
We calculate the mean of standard deviations of hand vertices $\sigma$ from 100 generations per object in the object/hand-centric frame on HO3D in Fig.~\ref{fig:diversity}.
All methods show comparable diversity in the object-centric frame but both methods can improve on the diversity of finger articulation.  
Note that standard deviation on its own is not a good metric as diverse samples may be implausible or ignore object affordance as visualized. 

\vspace{-1.2em}\paragraph{Grasp Characteristic. }
Fig.~\ref{fig:contactmap} visualizes the overall contact probability on hand across all generated grasps. The contact region of GraspTTA is centered at fingertips and (implausibly) even at the nail region shown on the back of the hand. Contact regions from \ours~are distributed on both fingers and palm, which is more consistent with how humans use their hands~\cite{brahmbhatt2019contactgrasp}.   

Tab.~\ref{tab:grasp} also reflects the same characteristics. Although \ours~has higher intersection volume, it has lowest average intersection depth and largest contact area. It also achieves the best performance in terms of grasp stability on HO3D and comparable results on out-of-domain \obman~objects. In user studies, \ours~is preferred against all methods on both datasets, even when comparing with ground-truth.

\begin{table}[t]
    \centering
\caption{\textbf{Ranking Grasps: } plausibility on HO3D over all grasps, along with the top and bottom 10\% grasps ranked by ~\ours. }
\label{tab:rank}
\vspace{-.5em}
\setlength{\tabcolsep}{2pt}
\resizebox{\linewidth}{!}{
\begin{tabular}{l l c c c c c c c c }
\toprule
& maxD$\downarrow$  & avgD$\downarrow$ & vol$\downarrow$ & disp $\downarrow$  & ratio$\uparrow$ & area$\uparrow$ \\
\cmidrule(lr){1-1}\cmidrule(lr){2-4} \cmidrule(lr){5-5} \cmidrule(lr){6-7}
\ours~(top 10\%) & \textbf{1.74} &  \textbf{0.31} & \textbf{10.57} &  \textbf{0.71} &  {1.00} &  {0.22} \\
\ours~(all) &  1.84 &  {0.31} & 11.46 &  {0.95} &  {1.00} &  {0.23} \\
\ours~(bottom 10\%) &  {1.87} &  0.33 & 13.11 &  1.41 &  {1.00} &  {0.23} \\
\bottomrule
\end{tabular}
}
\end{table}

\begin{figure}
    \centering
    \vspace{-1.5em}
    \includegraphics[width=\linewidth]{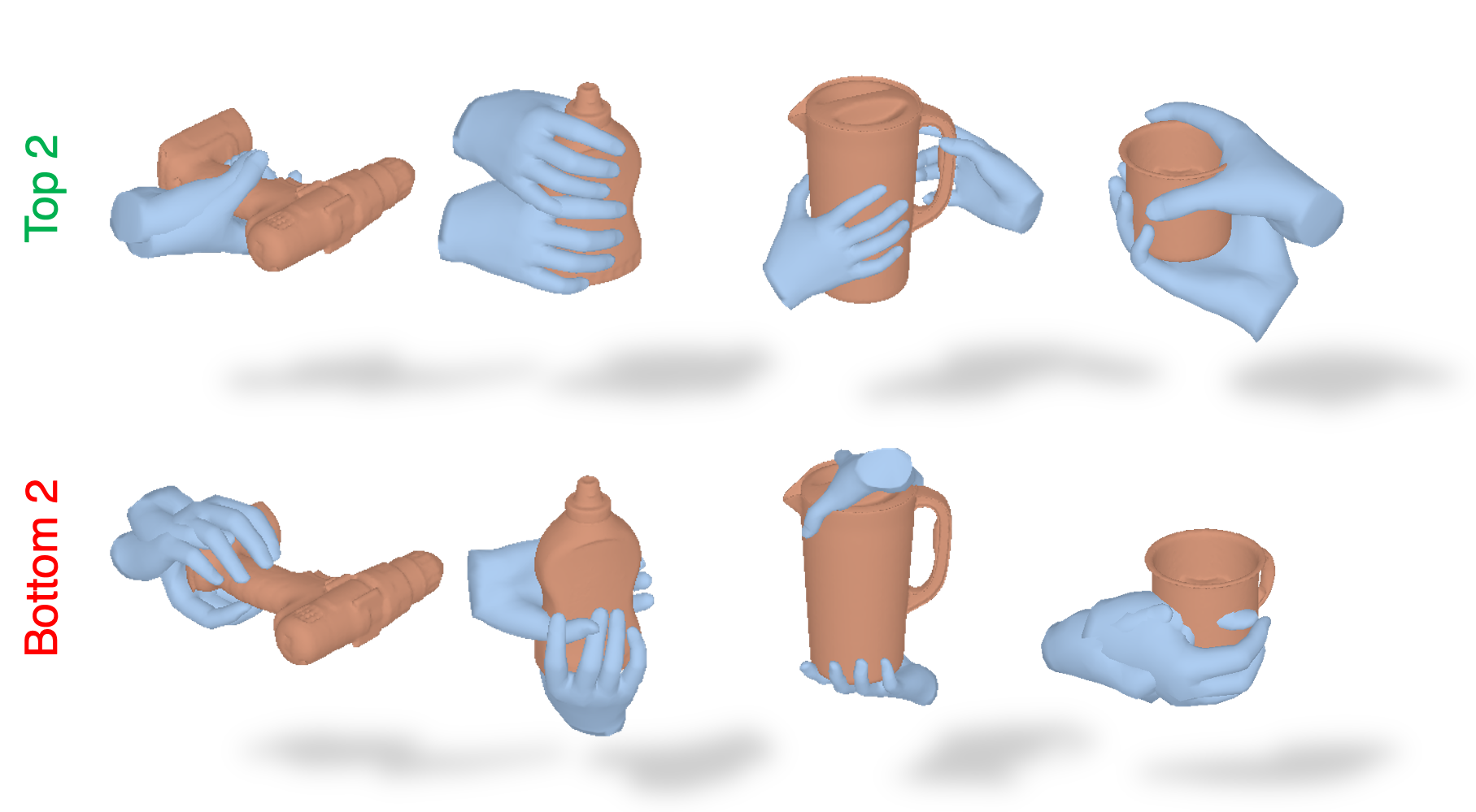}
    \vspace{-2em}
    \caption{\textbf{Ranking Grasps:} We visualize grasps with two highest scores (top) and two lowest scores (bottom) among 100 generated grasps from \ours.   }
    \label{fig:rank}
\vspace{-1em}
\end{figure}
\vspace{-1.2em}\paragraph{Ranking Grasps. }
Finally, we show that the proposed grasp score yields meaningful grasp ranking.   In Fig.~\ref{fig:rank}, we visualize top 2 and bottom 2 grasps out of 100 generations from \ours, evaluated by the proposed evaluation method. The ranking matches human's common sense. For example, power drills are often held in the middle; narrow side of bottles is often held upwards. Physically infeasible grasps are ranked low such as hands penetrating the mug.  Furthermore, the worst two grasps out of 100 are still reasonable in most cases. Note that all the grasps we show to users are randomly chosen for fair comparison. 
Quantitatively, top-ranked grasps in Tab.~\ref{tab:rank} show reduced simulation displacement and less intersection, validating our ranking approach's efficacy.

\section{Conclusion}
In this work, we propose a method to jointly generate 3D shape of HOI given an object category. Our method is the first to generate HOI across such diverse categories.
The learned prior G-HOP can serve as generic prior for relevant tasks like reconstructing interaction clips and human grasp synthesis, and we find that it leads to better performance than current task-specific baselines.  Despite the encouraging results, we are aware of several limitations: current method requires category information as input which may prevent the model from further scaling up;  there is no explicit mechanism to guarantee contact;  
and 
the model is still not at a scale comparable to generative models in other domains due to limited training data. Nevertheless, we believe that our work takes an encouraging step towards scaling up a general understanding of hand-object interactions. 

\vspace{-1em}
\paragraph{Acknowledgements.}
The authors would like to thank Hanwen Jiang and Korrawe Karunratanakul for clarifying baselines. 
We also thank  Fu-Jen Chu and Ruihan Gao for their detailed feedback on the manuscript. Yufei's PhD research is partially supported by a Google Gift. 

{
    \small
    \bibliographystyle{ieeenat_fullname}
    \bibliography{main}
}

\clearpage


\twocolumn[
\begin{@twocolumnfalse}
{
\begin{center}
\Large
\textbf{\ours: Generative Hand-Object  Prior \\for  Interaction Reconstruction and Grasp Synthesis} \\Supplementary Material \\\vspace{0.5em}
\end{center}
}
\end{@twocolumnfalse}
]
\appendix

In the supplementary materials, we provide more implementation details  and experimental results on the generative hand-object prior, prior-guided reconstruction, as well as  prior-guided grasp synthesis. We discuss network architecture (Sec.~\ref{sec:net}), effect of hand representation (Sec.~\ref{sec:field}), how to extract hand pose from skeletal distance field (Sec.~\ref{sec:sgd_pose}), and the text prompt we used (Sec.~\ref{sec:prompt}). Then, we show implementation details in reconstructing interaction clips and per-category results in Sec.~\ref{sec:supp_recon}. Furthermore, we analyze the effect of mesh refinement in grasp synthesis and discuss comparison with prior work Grasping Field~\cite{karunratanakul2020grasping} in Sec.~\ref{sec:supp_grasp}.

\section{Hand-Object Prior}


\subsection{Network Architecture}
\label{sec:net}
We use the same network architecture of latent autoencoder and 3D UNet diffusion model backbone as in SDFusion~\cite{cheng2023sdfusion}.  The 3D UNet backbone consists of several residual blocks. Each block is a stack of  GroupNorm layer~\cite{wu2018group}, non-linear activation~\cite{silu}, and 3D convolutional layer,  with optional cross attention layer to time embedding and text embedding. 
We provide an overview of network details and hyperparameters of our model in Tab.~\ref{tab:net_detail}. 

\begin{table}[ht]
    \centering
\begin{tabular}{lc}
\toprule
& {\ours} \\
\midrule
\( z \)-shape &  16$^3$ \( \times \) 3 \\
\(|\mathcal Z|\) & 8196 \\
Input Channel & 3 + 15 \\
Diffusion Steps & 1000 \\
Noise Schedule & linear \\
Channels & 64 \\
Number of Blocks & 3 \\
Attention resolutions & 4,2 \\ 
Channel Multiplier & 1,2,3 \\
Number of Heads & 8 \\
Transformers Depth & 1 \\
Batch Size & 64 \\
Iterations & 500k \\
Learning Rate & 1e-4 \\
\bottomrule
\end{tabular}
    \caption{\textbf{Network architecture for \ours.} } 
    \label{tab:net_detail}
\end{table}

\subsection{Ablating Skeletal Distance Field}
\label{sec:field}
\begin{figure}
    \centering
    \includegraphics[width=\linewidth]{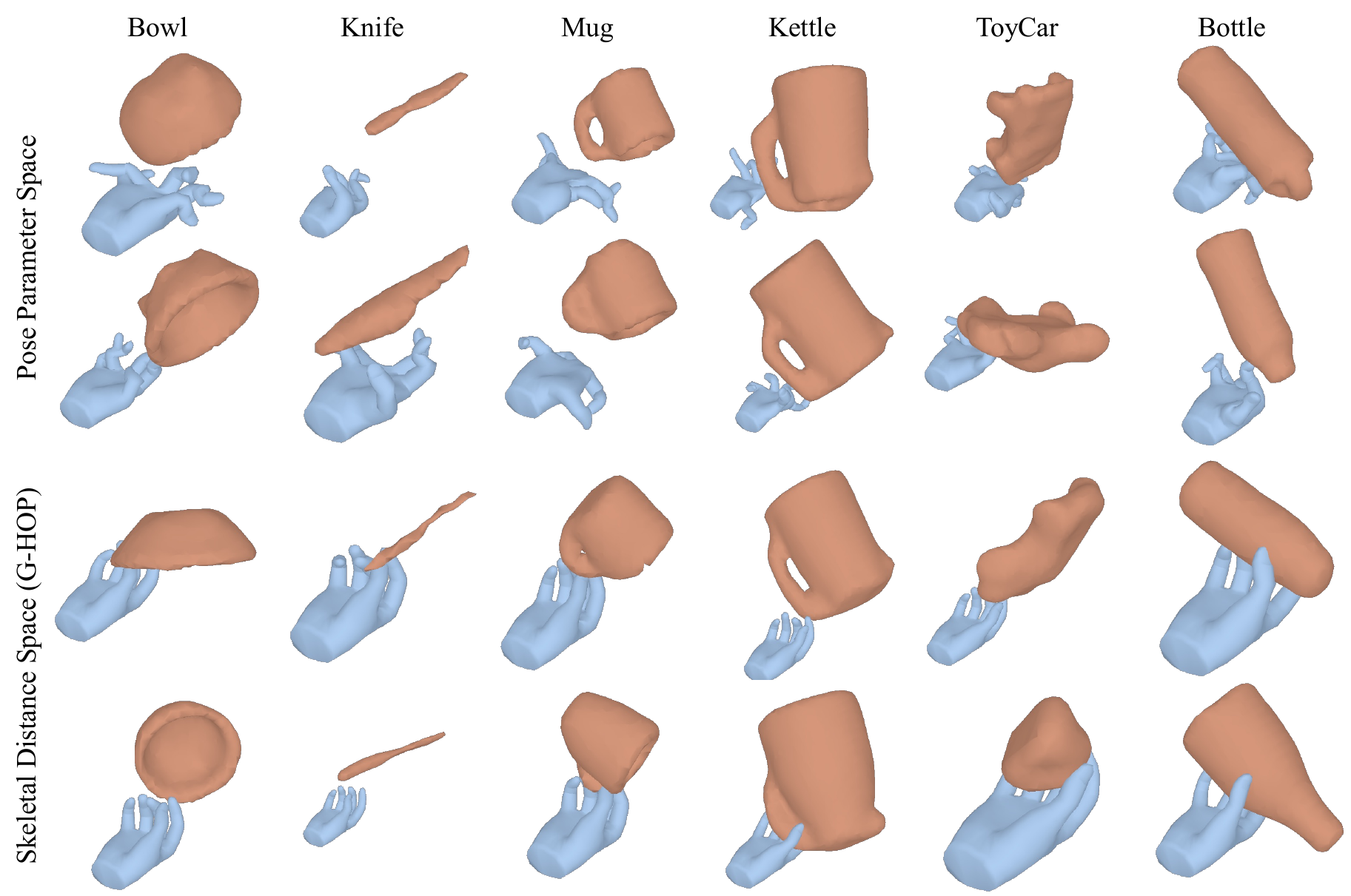}
    \caption{\textbf{Comparing Hand  Representation in Generative Hand-Object Prior:} Top 2 rows show the diffusion model that represents hand shape as pose parameters; bottom 2 rows show the diffusion model (ours) that represents hand shape as skeletal distance field. The homogeneous grid space is easier for the network to reason about interaction. }
    \label{fig:supp_prior}
\end{figure}

Many previous work~\cite{tevet2023human,karunratanakul2023guided} learn a diffusion model in the compact hand/human pose parameter space. We try to represent hand shape by hand pose parameters but find that this pose space is not optimal for jointly diffusing hand pose and objects in interaction. More specifically, the ablated method (pose parameter space) uses the same architecture as the main model except for the (noisy) pose parameter is passed via cross-attention layer instead of concatenating skeletal distance field to the object latent grid. 
We also search hyperparameters such as weights in DDPM loss to balance diffusing hand pose and diffusing object latent. We visualize the best ablated model in Fig.~\ref{fig:supp_prior} in comparison with our proposed model that represent hand shape by skeletal distance field. The diffusion model with pose parameter space struggles to generate plausible hand articulation together with objects. This is probably because the diffusion model is hard to reason about interaction in the heterogeneous space (1D hand pose and 3D object grids). 

\subsection{Hand Pose from Skeletal Distance Field}
\label{sec:sgd_pose}
Our proposed diffusion model generates skeletal distance field, from which hand pose parameters can be extracted. Given a target skeletal distance field $\hat H$, we optimize hand pose $\bm \theta$ such that its induced field is closer to the target, \ie $\bm \theta * = \arg \min_{\bm \theta} (H(\bm \theta) - \hat H)^2 + w \|\bm \theta\|_2^2$. We set $w$ to 1e-5 and optimizes for 1000 steps with Adam optimizer~\cite{adam} with learning rate 1e-2.  

\subsection{Text Prompt Template}
\label{sec:prompt}
We use the template ``a hand holding a \{\textit{category}\}" to convert category into a text prompt. 
In addition, we find that appending additional category attribute like size and shape beneficial when we scale up the number of category (see results in Sec.~\ref{sec:supp_recon}). It may be because attributes help to transfer information between categories with similar shapes but distinct semantics, \eg pens and spoons are all thin sticks.  We use LLM~\cite{gpt} to generate attribute automatically. We list text prompt we used in Tab.~\ref{tab:prompt}.


\definecolor{first}{rgb}{1.0, .83, 0.3}
\definecolor{second}{rgb}{1.0, 0.93, 0.7}
\def \first {\cellcolor{first}}
\def \second {\cellcolor{second}}
\def \third {}

\begin{table*}[t]
\footnotesize
\begin{center}
\caption{\textbf{Comparing Object Error of HOI Reconstruction on HOI4D.} }
\vspace{-1em}
\label{tab:supp_obj}

\setlength{\tabcolsep}{2pt}
\resizebox{\linewidth}{!}{
\begin{tabular}{l ccc ccc ccc ccc ccc ccc ccc}
\toprule

& \multicolumn{3}{c}{Mug} 
& \multicolumn{3}{c}{Bottle} 
& \multicolumn{3}{c}{Kettle} 
& \multicolumn{3}{c}{Bowl} 
& \multicolumn{3}{c}{Knife} 
& \multicolumn{3}{c}{ToyCar}  
& \multicolumn{3}{c}{\textbf{mean}} \\

\cmidrule(r){2-4} \cmidrule(r){5-7} \cmidrule(r){8-10} \cmidrule(r){11-13} \cmidrule(r){14-16} \cmidrule(r){17-19} \cmidrule(r){20-22}  
 
& F5$\uparrow$  & F10$\uparrow$ & CD$\downarrow$
& F5$\uparrow$  & F10$\uparrow$ & CD$\downarrow$ 
& F5$\uparrow$  & F10$\uparrow$ & CD$\downarrow$ 
& F5$\uparrow$  & F10$\uparrow$ & CD$\downarrow$ 
& F5$\uparrow$  & F10$\uparrow$ & CD$\downarrow$ 
& F5$\uparrow$  & F10$\uparrow$ & CD$\downarrow$ 
& F5$\uparrow$  & F10$\uparrow$ & CD$\downarrow$ \\

\midrule
iHOI~\cite{ye2022hand} & 0.44 & 0.71 & 2.1 & 0.47 & 0.77 & 1.5 & 0.21 & 0.45 & 6.3 & 0.38 & 0.64 & 3.1 & 0.33 & 0.68 & 2.8 & 0.66 & 0.95 & 0.5 & 0.42 & 0.70 & 2.7 \\
HHOR~\cite{huang2022hhor} & 0.18 & 0.37 & 6.9 & 0.26 & 0.56 & 3.1 & 0.12 & 0.30 & 11.3 & 0.31 & 0.54 & 4.2 & 0.71 & 0.93 & 0.6 & 0.26 & 0.59 & 1.9 & 0.31 & 0.55 & 4.7 \\
DiffHOI~\cite{ye2023vhoi} & 0.64 & 0.86 & 1.0 & 0.54 & 0.92 & 0.7 & 0.43 & 0.77 & 1.5 & 0.79 & 0.98 & 0.4 & 0.50 & 0.95 & 0.8 & 0.83 & 0.99 & 0.3 & 0.62 & 0.91 & 0.8 \\
\ours & 0.62 & 0.93 & 0.7 & 0.93 & 1.00 & 0.2 & 0.64 & 0.96 & 0.6 & 0.66 & 0.96 & 0.5 & 0.91 & 0.99 & 0.2 & 0.78 & 0.98 & 0.3 & 0.76 & 0.97 & 0.4 \\
\midrule
\ours(Cond) & 0.57 & 0.87 & 1.0 & 0.74 & 0.98 & 0.4 & 0.46 & 0.83 & 1.3 & 0.47 & 0.84 & 1.1 & 0.95 & 1.00 & 0.1 & 0.74 & 0.98 & 0.4 & 0.66 & 0.92 & 0.7 \\
\ours(2D) & 0.54 & 0.80 & 1.3 & 0.26 & 0.58 & 2.5 & 0.46 & 0.85 & 1.1 & 0.35 & 0.57 & 6.4 & 0.21 & 0.68 & 1.9 & 0.79 & 0.97 & 0.3 & 0.43 & 0.74 & 2.3 \\
\bottomrule
\end{tabular}
}
\end{center}
\end{table*}


\definecolor{first}{rgb}{1.0, .83, 0.3}
\definecolor{second}{rgb}{1.0, 0.93, 0.7}
\def \first {\cellcolor{first}}
\def \second {\cellcolor{second}}
\def \third {}

\begin{table*}[thp]
\footnotesize
\begin{center}
\caption{\textbf{Comparing Hand Error of HOI Reconstruction on HOI4D.}}
\vspace{-1em}
\label{tab:supp_hand}

\setlength{\tabcolsep}{2pt}
\resizebox{\linewidth}{!}{
\begin{tabular}{l cc cc cc cc cc cc cc}
\toprule

& \multicolumn{2}{c}{Mug} 
& \multicolumn{2}{c}{Bottle} 
& \multicolumn{2}{c}{Kettle} 
& \multicolumn{2}{c}{Bowl} 
& \multicolumn{2}{c}{Knife} 
& \multicolumn{2}{c}{ToyCar}  
& \multicolumn{2}{c}{\textbf{mean}} \\

\cmidrule(r){2-3} \cmidrule(r){4-5} \cmidrule(r){6-7} \cmidrule(r){8-9} \cmidrule(r){10-11} \cmidrule(r){12-13} \cmidrule(r){14-15}  
 
& MPJPE$\downarrow$  & AUC$\uparrow$
& MPJPE$\downarrow$  & AUC$\uparrow$
& MPJPE$\downarrow$  & AUC$\uparrow$
& MPJPE$\downarrow$  & AUC$\uparrow$
& MPJPE$\downarrow$  & AUC$\uparrow$
& MPJPE$\downarrow$  & AUC$\uparrow$
& MPJPE$\downarrow$  & AUC$\uparrow$ \\

\midrule
iHOI~\cite{ye2022hand} & 1.10 & 0.78 & 1.09 & 0.78 & 1.11 & 0.78 & 1.23 & 0.76 & 1.39 & 0.72 & 1.20 & 0.76 & 1.19 & 0.76 \\
DiffHOI~\cite{ye2023vhoi} & 1.06 & 0.79 & 1.01 & 0.80 & 1.07 & 0.79 & 1.21 & 0.76 & 1.33 & 0.73 & 1.04 & 0.79 & 1.12 & 0.78 \\
\ours  & 1.02 & 0.80 & 0.97 & 0.81 & 0.98 & 0.81 & 1.09 & 0.78 & 1.20 & 0.76 & 1.02 & 0.80 & 1.05 & 0.79 \\
\midrule
\ours(Cond) & 1.08 & 0.78 & 1.06 & 0.79 & 1.09 & 0.79 & 1.18 & 0.76 & 1.34 & 0.73 & 1.11 & 0.78 & 1.14 & 0.77 \\
\ours(2D) & 1.10 & 0.78 & 0.97 & 0.81 & 1.06 & 0.79 & 1.24 & 0.75 & 1.24 & 0.75 & 1.07 & 0.79 & 1.11 & 0.78 \\

\bottomrule
\end{tabular}
}
\end{center}
\end{table*}

\begin{table}[thp]
\footnotesize
\begin{center}
\caption{\textbf{Comparing Hand-Object Alignment (CD$_h \downarrow$) of HOI Reconstruction on HOI4D. } }
\vspace{-1em}
\label{tab:supp_align}

\setlength{\tabcolsep}{2pt}
\resizebox{\linewidth}{!}{
\begin{tabular}{l c c c c c c c}
\toprule

& \multicolumn{1}{c}{Mug} 
& \multicolumn{1}{c}{Bottle} 
& \multicolumn{1}{c}{Kettle} 
& \multicolumn{1}{c}{Bowl} 
& \multicolumn{1}{c}{Knife} 
& \multicolumn{1}{c}{ToyCar}  
& \multicolumn{1}{c}{\textbf{mean}} \\

 

\midrule
iHOI~\cite{ye2022hand} & 19.7 & 13.9 & 35.9 & 49.3 & 21.9 & 21.6 & 27.1 \\
HHOR~\cite{huang2022hhor} & 229.1 & 172.0 & 100.4 & 50.1 & 185.1 & 255.8 & 165.4 \\
DiffHOI~\cite{ye2023vhoi} & 18.1 & 15.3 & 42.2 & 101.8 & 91.6 & 23.3 & 48.7 \\
\ours  & 12.4 & 9.7 & 41.8 & 26.2 & 13.2 & 7.5 & 18.4 \\
\midrule
\ours(Cond) & 10.2 & 6.9 & 40.7 & 10.4 & 39.1 & 8.5 & 19.3 \\
\ours(2D) & 14.5 & 33.6 & 61.6 & 71.0 & 141.7 & 38.8 & 60.2 \\
\bottomrule
\end{tabular}
}

\end{center}
\end{table}


\definecolor{first}{rgb}{1.0, .83, 0.3}
\definecolor{second}{rgb}{1.0, 0.93, 0.7}
\def \first {\cellcolor{first}}
\def \second {\cellcolor{second}}
\def \third {}

\begin{table}[t]
\footnotesize
\begin{center}
\caption{\textbf{Additional Ablation Studies of HOI reconstruction:} We report object error (F@5mm, F@10mm, CD), hand-object alignment CD$_h$, and hand error (MPJPE, AUC) on HOI4D. We analyze the effect of other implementation details, including dynamic noise thresholding and choice of text prompt templates. }
\vspace{-1em}
\label{tab:recon_ablation}

\setlength{\tabcolsep}{2pt}
\resizebox{\linewidth}{!}{
\begin{tabular}{l c c c c c c c}
\toprule

& \multicolumn{3}{c}{Object Error} 
& \multicolumn{1}{c}{Align}
& \multicolumn{2}{c}{Hand Error} \\

\cmidrule(r){2-4} \cmidrule(r){5-5} \cmidrule(r){6-7}

& F5$\uparrow$  & F10$\uparrow$ & CD$\downarrow$  & CD$_h$ $\downarrow$ & MPJPE$\downarrow$ & AUC$\uparrow$ \\

\midrule

\ours & 0.76 & 0.97 & 0.4 & 18.4 & 1.05 & 0.79 \\
$U_b=0.25$ & 0.69 & 0.95 & 0.5 & 50.0 & 1.01 & 0.80 \\
$U_b=0.75$ & 0.49 & 0.76 & 4.0 & 48.1 & 1.06 & 0.79 \\
\midrule
\ours~(G) w/ attr & 0.65 & 0.92 & 0.7 & 17.8 & 1.06 & 0.79 \\
\ours~(G) wo/ attr & 0.61 & 0.89 & 0.8 & 24.6 & 1.04 & 0.79 \\
\bottomrule
\end{tabular}
}
\end{center}
\vspace{-3.5em}
\end{table}

\section{Reconstructing Interaction Clips}
\label{sec:supp_recon}
Following prior work~\cite{ye2023vhoi}, we evaluate reconstruction on two sequences per category on HOI4D. We report mean performance per category in terms of object error (Tab.~\ref{tab:supp_obj}), hand error (Tab.~\ref{tab:supp_hand}), and their alignment (Tab.~\ref{tab:supp_align}). 
 In addition to baselines and ablations reported in main paper, we also analyze the effect of other implementation details as follows:

\paragraph{Dynamic Noise Threshold. } The amount of injected noise in SDS has large impact on the guidance effect. We find that thin structures are better captured when adding a smaller noise while thick structure are better captured when adding larger noise. We use an adaptive noise scheduler that dynamically adjusts the maximum amount of noise $U_b, i \sim \mathcal{U}[U_a,U_b]$ based on the current object shape. More specifically, it is a linear interpolation based on minimal object SDF value in the current representation, \ie
\begin{align*}
U_b &= \frac{s - s_{\min}} {s_{\max} - s_{\min}} U_{b{\max}} + (1-\frac{s - s_{\min}} {s_{\max} - s_{\min}}) U_{b{\min}} \\ s &= clamp(\min O[X_{grid}], s_{\min}, s_{\max})    
\end{align*}
In our experiment, we set $U_{b\max} = 0.75, U_{b\min} = 0.25, s_{\min}=-0.2, s_{\max} = -0.01$. As reported in Tab.~\ref{tab:recon_ablation}, our dynamic noise threshold leads to better performance than constant noise threshold. 

\paragraph{Scaling Up Number of Categories. } For fair comparison, we use the diffusion model that only trains on HOI4D dataset to reconstruct interaction clips. In Tab.~\ref{tab:recon_ablation},, we also compare with the generalist model (\ours~(G) ) that trains on all seven datasets. Note that we use \ours~(G) in all other experiments. We find that adding attribute to text prompt helps when scaling up to more categories.  
While \ours~(G) leads to a bit worse reconstruction performance on the HOI4D dataset than the specialist which is trained only on HOI4D,  it still outperforms other baselines. 

\begin{figure}
    \centering
    \includegraphics[width=\linewidth]{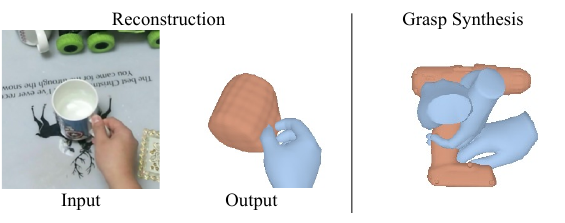}
    \caption{\textbf{2D Joint Prior (DiffHOI-J):} reconstruction and grasp
synthesis results guided by 2D joint prior. }
    \label{fig:2djoint}
\end{figure}
\paragraph{2D Joint Prior.} We trained a joint prior version of DiffHOI, or a 2D version of \ours ~$p(\pi(O), \pi(H) | C)$. Interestingly, we find that this cannot effectively guide grasp synthesis or reconstruction (Fig.~\ref{fig:2djoint}, Tab.~\ref{tab:supp_obj}-\ref{tab:supp_align}). It performs even worse than DiffHOI~\cite{ye2023vhoi}, perhaps because it is  harder to learn the distribution over object, hand,  and rendering viewpoints (unlike DiffHOI where the `conditioning' informs about the hand and viewpoint).

\section{Grasp Synthesis}
\label{sec:supp_grasp}

\begin{table}[t]
\footnotesize
\begin{center}
\caption{\textbf{Comparison with Baselines:} We compare human grasp synthesis along with prior work GF~\cite{karunratanakul2020grasping}. * denotes GF's evaluation setting with known object pose. }
\label{tab:supp_grasp}

\setlength{\tabcolsep}{2pt}
\resizebox{\linewidth}{!}{
\begin{tabular}{l c c c c c c c c c }
\toprule

&
& \multicolumn{3}{c}{Intersection} 
& \multicolumn{1}{c}{Disp.}
& \multicolumn{2}{c}{Contact}
\\
\cmidrule(lr){3-5} \cmidrule(lr){6-6} \cmidrule(lr){7-8}
& & max D $\downarrow$  & avg D  $\downarrow$ & vol $\downarrow$ & avg $\downarrow$  & ratio $\uparrow$ & area $\uparrow$ \\
\midrule
\multirow{5}{*}{{ObMan}}

& GF~\cite{karunratanakul2020grasping}* & 0.56 & 0.44 & 6.05 & 2.07 & 0.89 & 0.06 \\
& \ours* & 0.97 & 0.70 & 6.39 & 2.03 & 1.00 & 0.13 \\
\cmidrule{2-8}
& GF~\cite{karunratanakul2020grasping} & 0.79 & 0.64 & 43.35 & 1.82 & 1.00 & 0.09 \\
& \ours & 0.74 & 0.51 & 17.40 & 1.85 & 0.93 & 0.25 \\
\bottomrule
\end{tabular}
}

\end{center}
\end{table}

\begin{table}[t]
\footnotesize
\begin{center}
\caption{\textbf{Effect of Refinement:} We report human grasp synthesis before and after mesh refinement. \ours$\dagger$ denotes generated grasps before mesh refinement. }
\label{tab:supp_refine}

\setlength{\tabcolsep}{2pt}
\resizebox{\linewidth}{!}{
\begin{tabular}{l c c c c c c c c c }
\toprule

&
& \multicolumn{3}{c}{Intersection} 
& \multicolumn{1}{c}{Disp.}
& \multicolumn{2}{c}{Contact}
\\
\cmidrule(lr){3-5} \cmidrule(lr){6-6} \cmidrule(lr){7-8}
& & max D $\downarrow$  & avg D  $\downarrow$ & vol $\downarrow$ & avg $\downarrow$  & ratio $\uparrow$ & area $\uparrow$ \\
\midrule
\multirow{2}{*}{{ObMan}}
& \ours$\dagger$ & 0.74 & 0.57 & 8.25 & 3.87 & 0.82 & 0.12 \\
& \ours & 0.74 & 0.51 & 17.40 & 1.85 & 0.93 & 0.25 \\
\midrule
\multirow{2}{*}{{HO3D}}
& \ours$\dagger$ & 1.84 & 0.31 & 11.46 & 0.95 & 1.00 & 0.23 \\
& \ours & 2.42 & 0.68 & 7.55 & 2.48 & 0.99 & 0.20 \\\bottomrule
\end{tabular}
}

\end{center}
\end{table}

\begin{figure}
    \centering
    \includegraphics[width=\linewidth]{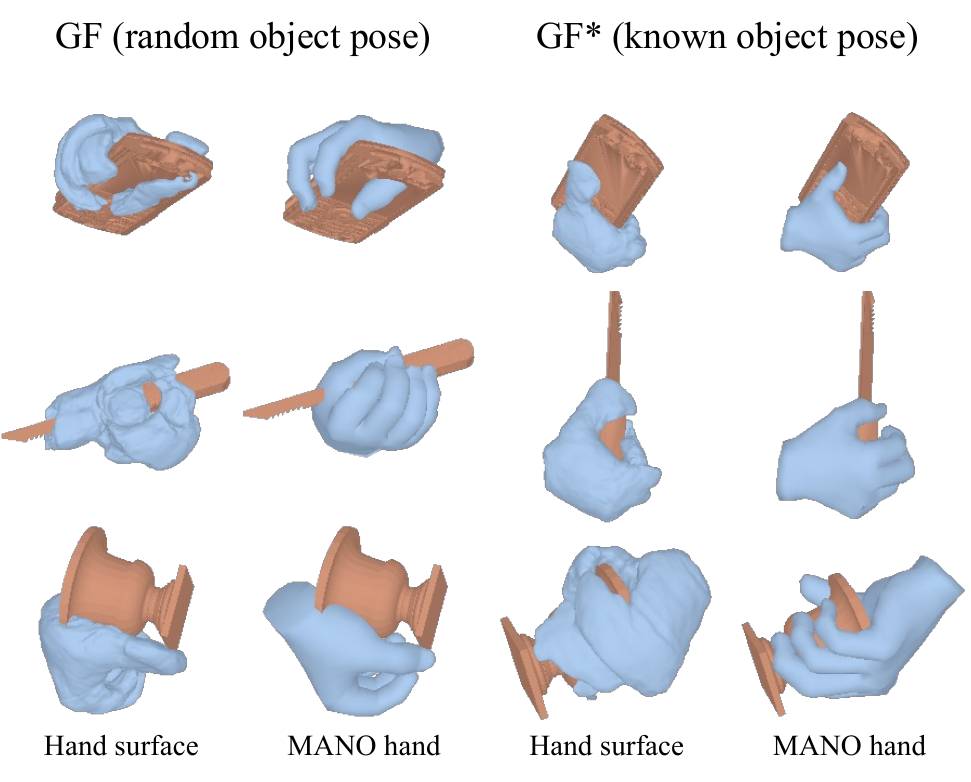}
    \caption{GF assumes known object pose when evaluating. Randomizing object pose affects their performance. }
    \label{fig:gf}
\end{figure}

\paragraph{Comparison with Grasping Field. }
Grasping Field~\cite{karunratanakul2020grasping} is a representative method that uses a conditional VAE to generate hand surface distance field given an object point cloud. Their evaluation setup generates grasps for known object pose with respect to hand. We evaluate \ours~ under their setup by only optimizing hand articulation while keeping the relative pose as the given ground truth. We denote this setting with known object pose as *. \ours~  also benefits from well initialized object poses as contact ratio increases to 100\%. Our  contact area reduces probably because the given object pose are obtained from GT grasps that uses more finger tips and this makes the human hand palm harder to make contact. 
 We also show that randomizing the relative pose (our evaluation setup) significant affects their performance, as visualized in Fig.~\ref{fig:gf}. 
Note that GF gets large intersection volume but less intersection depth. This is because the latter is only calculated on each hand vertices  inside of the object. For example, in the second row of Fig.~\ref{fig:gf}, the knife penetrates hand, leading to high volume. But the maximum intersection depth for each hand vertices is less than the thickness of the knife.

\paragraph{Effect of Refinement}
After optimizing human grasps with respect to SDS loss using object SDF grid, we also do a light-weight mesh refinement by replacing the object SDF grid with the original mesh. It is to account for loss of accuracy during mesh conversion. We use the same objectives in previous work~\cite{hasson19_obman,ye2022hand} that encourage contact and discourage penetration. We denote the generated grasps before mesh refinement as $\dagger$ and report its performance  on two datasets in Tab.~\ref{tab:supp_refine}. Even without mesh refinement, the generated grasps also have large contact area and less displacement in simulation. The refinement process can adjust hand pose to further improves the contact and grasp stability. 

\begin{figure}
    \centering
    \includegraphics[width=\linewidth]{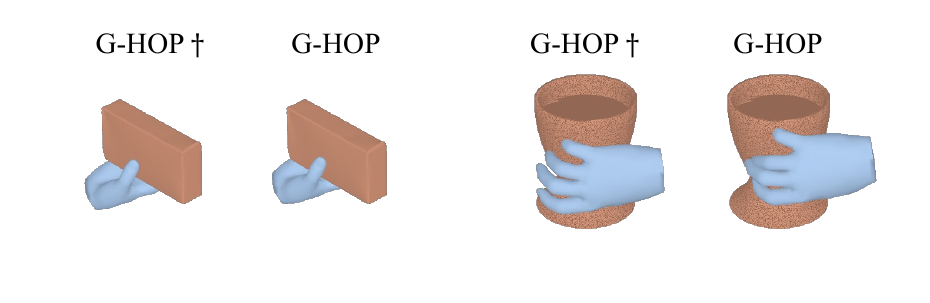}
    \caption{\textbf{Effect of Mesh Refinement: } We visualize synthesized grasps before (\ours$\dagger$) and after (\ours) refinement. }
    \label{fig:refine}
\end{figure}

\begin{figure}
    \centering
    \includegraphics[width=\linewidth]{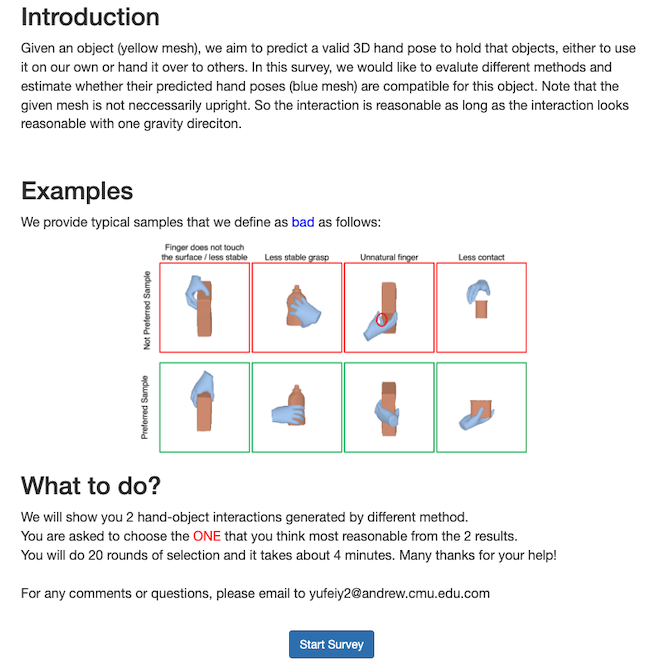}
    \includegraphics[width=\linewidth]{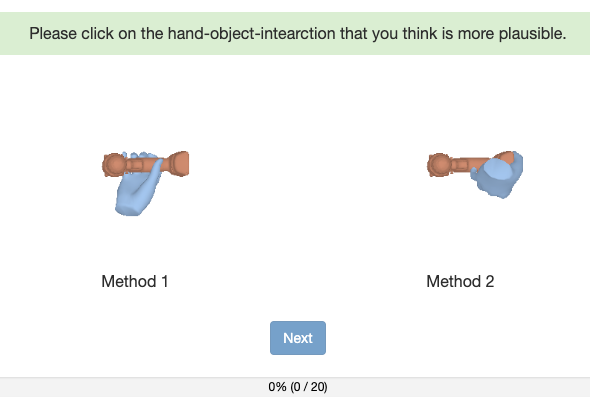}
    \caption{\textbf{User Study Interface: } We visualize user study interface including the user instruction page and the survey page. }
    \label{fig:user_interface}
\end{figure}

\subsection{User Study Interface}
Fig.~\ref{fig:user_interface} shows the user interface for evaluating the generated grasps. Users are presented two grasps visualized from different view angles as gif and are asked to choose the more plausible grasps.

\clearpage

\tablefirsthead{\toprule Class&\multicolumn{1}{c}{Attribute} \\ \midrule}
\tablehead{%
\multicolumn{2}{c}%
{{\bfseries  Continued from previous column}} \\
\toprule
Class&\multicolumn{1}{c}{Attribute}\\ \midrule}
\tabletail{%
\midrule \multicolumn{2}{r}{{Continued on next column}} \\ \midrule}
\tablelasttail{%
\\\midrule
\multicolumn{2}{r}{{Concluded}} \\ \bottomrule}
\footnotesize
    \topcaption{We provide list of class names and their attributes used in the text prompt. The class names are manually merged across different datasets while the attributes are automatically generated by large language model~\cite{gpt}.}
    \label{tab:prompt}
\begin{supertabular}{p{0.2\linewidth} p{0.69\linewidth}}
    plate& medium, flat, circular\\
    baseboard& big, long, rectangular\\
    stamp& small, flat, square\\
    laptop& big, flat, rectangular\\
    funnel& medium, conical\\
    spatula& medium, flat, elongated\\
    pear& small, pear shaped\\
    lemon& small, oval\\
    stick& varies, cylindrical, long\\
    cylinder& varies, cylindrical\\
    mug& medium, cylindrical, handle attached\\
    flute& medium, cylindrical, long\\
    shield& big, curved, oval or round\\
    floor& big, flat, rectangular or irregular\\
    mouse& medium, oval, handheld\\
    fish& varies, animal shaped\\
    screw driver& medium, cylindrical, elongated\\
    pen& small, cylindrical, elongated\\
    hair dryer& medium, elongated, handheld\\
    burger& medium, cylindrical, stacked layers\\
    paint roller& medium, cylindrical, handheld\\
    power saw& big, elongated, handheld or standalone\\
    bottle& medium, cylindrical, narrow neck\\
    pump& varies, mechanical, various shapes\\
    flask& medium, cylindrical or conical, narrow neck\\
    sheet& big, flat, rectangular\\
    hand bag& medium, varies, handle attached\\
    stapler& medium, rectangular, handheld\\
    gummy& small, animal or object shaped\\
    fork& small, elongated, tines at one end\\
    wood& varies, solid, various shapes\\
    chopsticks& small, cylindrical, elongated\\
    strawberry& small, heart-shaped\\
    cupmod& medium, cylindrical, handle attached\\
    spray& medium, cylindrical, nozzle at top\\
    crate& big, cuboid, open structure\\
    microwave& big, rectangular, box-like\\
    headphone& medium, round or oval, worn over ears\\
    apple& small, round, stem at top\\
    backpack& big, varies, straps attached\\
    brick& medium, rectangular, solid\\
    wood plank& big, flat, rectangular\\
    tv& big, flat, rectangular\\
    rubiks& small, cubical, multicolored faces\\
    carpet& big, flat, rectangular or oval\\
    container& varies, solid, various shapes\\
    lego& small, rectangular or square, connecting knobs\\
    jar& medium, cylindrical or oval, lid on top\\
    oven& big, box-like, door at front\\
    mixer& big, varies, mechanical\\
    train& big, cylindrical, long\\
    teddy bear& medium, animal shaped, soft\\
    chess rook& small, cylindrical, castle-shaped top\\
    binoculars& medium, cylindrical, two lenses\\
    pencil mod& small, cylindrical, elongated\\
    knife& medium, flat, sharp edge\\
    tin& medium, cylindrical or rectangular, lid on top\\
    light tube& medium, cylindrical, elongated\\
    ball& small, spherical\\
    cupcake& small, cylindrical, rounded top\\
    spoon& small, oval or round, handle attached\\
    chalk& small, cylindrical, elongated\\
    light bulb& small, round, screw base\\
    case& varies, box-like, lid or zipper\\
    peg test& varies, varies, testing equipment\\
    piggy bank& medium, animal shaped, slot on top\\
    kettle& medium, rounded, spout and handle\\
    wrench& medium, elongated, adjustable jaw\\
    bacon& small, flat, elongated\\
    purse& medium, varies, handle or strap\\
    boat& big, elongated, hollow\\
    disk& small, flat, circular\\
    game controller& medium, ergonomic, buttons and joysticks\\
    keyboard& medium, flat, rectangular\\
    trowels& medium, flat, handle attached\\
    shovel& big, flat, long handle\\
    eye glasses& small, oval or round, frame with lenses\\
    stanford bunny& small, animal shaped, 3D model\\
    camera& medium, box-like, lens at front\\
    rifle& big, elongated, barrel and stock\\
    can& small, cylindrical, lid on top\\
    range& big, flat or box-like, knobs and burners\\
    toy airplane& small, aerodynamic, wings attached\\
    cube& varies, cubical\\
    tablet& medium, flat, rectangular\\
    teapot& medium, rounded, spout and handle\\
    chair& big, varies, seat and backrest\\
    beaker& small, cylindrical, pouring lip\\
    plum& small, round, pit inside\\
    triangle& varies, triangular\\
    barrel& big, cylindrical, hollow\\
    cup& small, cylindrical, handle attached\\
    toothpaste& small, cylindrical, tube-shaped\\
    bag& varies, varies, handle or strap\\
    pyramid& varies, pyramidal\\
    dice& small, cubical, numbered faces\\
    ruler& small, flat, rectangular\\
    scissors& small, paired blades, handles\\
    clamp& small, C or G shaped, screw mechanism\\
    phone& medium, flat, rectangular\\
    marbles& small, spherical, glass or clay\\
    dart& small, conical, pointed tip\\
    calculator& medium, flat, rectangular\\
    duck& varies, animal shaped\\
    chain& varies, interlinked, metal\\
    bucket& medium, cylindrical, handle attached\\
    peach& small, round, pit inside\\
    donut& small, cylindrical, hole in center\\
    flashlight& medium, cylindrical, light at one end\\
    sponge& small, soft, varies\\
    mat& medium, flat, rectangular or oval\\
    cardboard& varies, flat, rectangular\\
    scoop& small, semi-spherical, handle attached\\
    block& varies, solid, cuboidal\\
    pliers& medium, paired jaws, handles\\
    board& big, flat, rectangular\\
    shoe& medium, foot-shaped, footwear\\
    floor mate& varies, flat, used for cleaning\\
    brush& varies, bristles attached, handle\\
    alarm clock& small, circular or square, time display\\
    hood& big, curved, worn over head\\
    pot& medium, cylindrical, handle attached\\
    chessboard& medium, square, 8x8 squares\\
    pillow& medium, soft, rectangular\\
    power drill& medium, cylindrical, elongated\\
    marshmallow& small, cylindrical or cubic, soft\\
    bowl& medium, round, hollow\\
    tube& varies, cylindrical, hollow\\
    frisbee& medium, flat, circular\\
    hammer& medium, heavy head, handle attached\\
    toothbrush& small, bristles at end, handle\\
    toycar& small, car shaped, wheels attached\\
    elephant& big, animal shaped\\
    tray& medium, flat, raised edges\\
    box& varies, cuboidal, lid or flaps\\
    book& medium, flat, rectangular\\
    skillet lid& medium, flat or domed, handle on top\\
    table& big, flat, supported by legs\\
    banana& small, curved, elongated\\
    padlock& small, rounded or square, shackle on top\\
    bin& big, cylindrical or cuboidal, open top\\
    blender& medium, cylindrical, mechanical\\
    pitcher& medium, cylindrical, handle and spout\\
    toilet& big, bowl-shaped, plumbing fixture\\
    wine glass& small, stemmed, conical\\
    towel& big, flat, rectangular\\
    vacuum& big, cylindrical, mechanical\\
    chips& small, flat, round or oval\\
    orange& small, round, citrus fruit\\
    microphone& small, cylindrical, handheld\\
    usb stick& small, rectangular, electronic\\
    door knob& small, round, mounted on door\\
    fryingpan& medium, flat, round\\
    watch& small, round, straps attached\\
    eraser& small, rectangular or cylindrical, soft \\
\end{supertabular}%

\end{document}